\documentclass[10pt,twocolumn,letterpaper]{article}

\pdfoutput=1

\usepackage{cvpr}

\usepackage[T1]{fontenc}
\usepackage[utf8]{inputenc}
\usepackage[english]{babel}
\usepackage{newtxtext}
\usepackage[vvarbb]{newtxmath}
\usepackage{courier}
\usepackage{textcomp}
\usepackage{microtype}
\usepackage{hyphenat}
\usepackage[all]{nowidow}

\usepackage{amsmath}
\usepackage{amssymb}
\usepackage{amsfonts}
\usepackage{xfrac}

\usepackage{booktabs}
\usepackage{tabulary}

\usepackage{epsfig}
\usepackage{graphicx}
\usepackage[labelsep=period,labelfont=bf,font={small}]{caption}
\usepackage{subcaption}
\usepackage{dblfloatfix} 
\usepackage{pifont}

\usepackage{xcolor}
\usepackage{soul}

\usepackage[pagebackref=true,breaklinks=true,letterpaper=true,colorlinks,bookmarks=false]{hyperref}

\hypersetup{pdfcreator={Tom F.H. Runia, Cees G.M. Snoek and Arnold W.M. Smeulders}}
\hypersetup{pdfproducer={Tom F.H. Runia}}
\hypersetup{pdftitle={Real-World Repetition Estimation by Div, Grad and Curl}}
\hypersetup{pdfsubject={Computer Vision}}
\hypersetup{pdfkeywords={vision, video, flow, motion, divergence, gradient, curl, grad, periodicity, recurrence, repetition, counting, actions}}

\graphicspath{{./images/}}

\newcommand{\comment}[1]{\ignorespaces}

\newcommand{\datasetname}{\emph{QUVA Repetition}}
\newcommand{\ytsegments}{\emph{YTSegments}}
\newcommand{\datasetnamebold}{\textbf{QUVA Repetition}}
\newcommand{\ytsegmentsbold}{\textbf{YTSegments}}

\def\*#1{\mathbf{#1}}  
\DeclareMathAlphabet\mathcalbf{OMS}{cmsy}{b}{n}
\newcommand{\nablabf}{\boldsymbol{\nabla}}

\newcommand{\grad}{\nablabf \*F}
\newcommand{\divergence}{\nablabf \boldsymbol{\cdot} \*F}
\newcommand{\curl}{\nablabf \boldsymbol{\times} \*F}

\newcommand{\gradd}{\nablabf \mathcalbf{F}}
\newcommand{\divergencee}{\nablabf \boldsymbol{\cdot} \mathcalbf{F}}
\newcommand{\curll}{\nablabf \boldsymbol{\times} \mathcalbf{F}}

\AtBeginDocument{
 \abovedisplayskip=10pt plus 3pt minus 6pt
 \abovedisplayshortskip=0pt plus 3pt
 \belowdisplayskip=10pt plus 3pt minus 6pt
 \belowdisplayshortskip=7pt plus 3pt minus 4pt
}

\definecolor{highlightcolor}{RGB}{255,237,191}
\definecolor{textboxcolor}{RGB}{155,155,155}
\definecolor{paperboxcolor}{RGB}{204,148,73}
\definecolor{highlightred}{RGB}{224,103,103}

\cvprfinalcopy 

\def\cvprPaperID{103} 

\title{Real-World Repetition Estimation by Div, Grad and Curl}

\author{Tom F.~H. Runia \hspace{1.7em} Cees G.~M. Snoek \hspace{1.4em} Arnold W.~M. Smeulders \\
    QUVA Deep Vision Lab, University of Amsterdam\\
    {\tt\small \{runia,cgmsnoek,a.w.m.smeulders\}@uva.nl}
}

\begin{document}

\maketitle

\begin{abstract}
  We consider the problem of estimating repetition in video, such as performing push-ups, cutting a melon or playing violin. Existing work shows good results under the assumption of static and stationary periodicity. As realistic video is rarely perfectly static and stationary, the often preferred Fourier-based measurements is inapt. Instead, we adopt the wavelet transform to better handle non-static and non-stationary video dynamics. From the flow field and its differentials, we derive three fundamental motion types and three motion continuities of intrinsic periodicity in 3D. On top of this, the 2D perception of 3D periodicity considers two extreme viewpoints. What follows are $18$ fundamental cases of recurrent perception in 2D. In practice, to deal with the variety of repetitive appearance, our theory implies measuring time-varying flow $\mathbf{F}_t$ and its differentials $\grad_t$, $\divergence_t$ and $\curl_t$ over segmented foreground motion. For experiments, we introduce the new QUVA Repetition dataset, reflecting reality by including non-static and non-stationary videos. On the task of counting repetitions in video, we obtain favorable results compared to a deep learning alternative.
\end{abstract}



\vspace{-0.2cm}

\section{Introduction}
\label{sec:introduction}

Visual repetition is ubiquitous in the world around us. It is present in activities like rowing, music-making and cooking. It arises in natural and urban environments: traffic patterns, blinking lights, and leaves in the wind. Rhythm and repetition are used to approximate velocity, estimate progress and to trigger attention \cite{johansson1973visual}. In computer vision, understanding repetition in video is important as it can serve action classification \cite{goldenberg2005behavior,lu2004repetitive}, action localization \cite{laptev2005periodic,sarel2005separating}, human motion analysis \cite{albu2008generic,ran2007pedestrian}, 3D reconstruction \cite{belongie2006structure} and camera calibration \cite{huang2016camera}. Estimating repetition remains challenging. First and foremost, repetition appears in many forms due to its variety in motion pattern and motion continuity. The viewpoint is crucial for the perception of recurrence. In practice, camera motion makes repetition estimation inevitably hard.

\begin{figure}[t]
  \centering
  \includegraphics[width=\columnwidth]{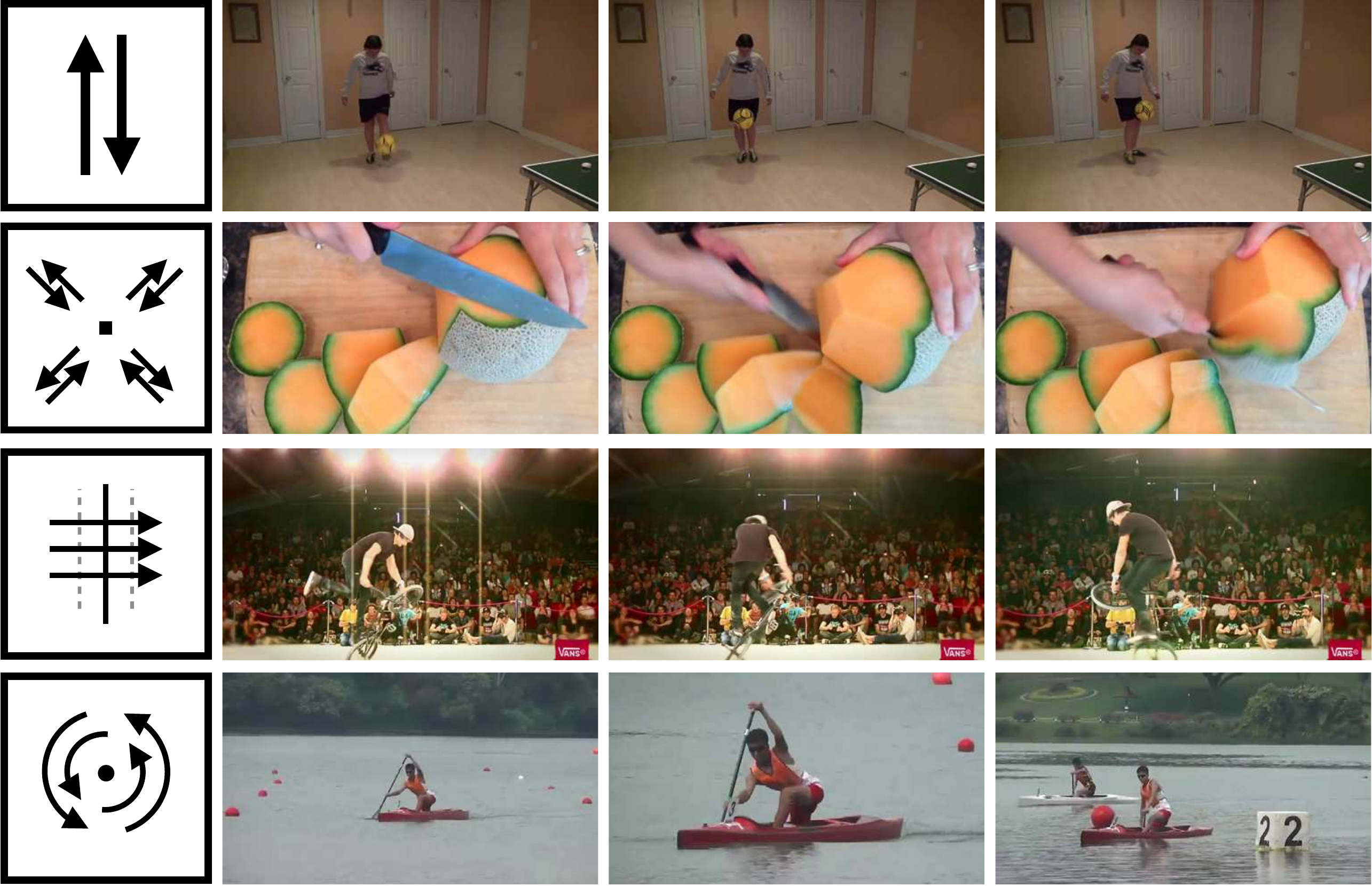}
  \caption{Four examples of visual repetition under realistic circumstances. The first two rows show \emph{oscillatory translation} under two different viewpoints. Similarly for \emph{constant rotation} in the bottom rows. The abstraction on the left symbolizes the perceived flow in 2D, to be detailed in Section~\ref{sec:recurrence-in-video}.
  \label{fig:intro-figure}}
  \vspace{-0.4cm}
\end{figure}

Existing work on repetition estimation in video \cite{levy2015live,pogalin2008visual} reports good results under the assumption that the motion is well-localized (static) and strongly periodic (stationary). In short, existing work focuses on video that is static in every aspect of repetition. As real life is more complex, our method relies on motion foreground segmentation to localize the salient motion and handle non-static video. Furthermore, we found fixed-period Fourier analysis \cite{cutler2000robust,pogalin2008visual,polana1997detection} to be unsuitable for repetition estimation in real-world video as non-stationarity often appears. To permit non-stationary video dynamics, we adopt the wavelet transform for decomposing video signals into a time-frequency spectrum.

We reconsider the theory of repetition \cite{pogalin2008visual,davis2000categorical} starting from the divergence, gradient and curl operators acting on the 3D flow field. We derive three \emph{motion types} and three \emph{motion continuities}. What follows are $3\times3$ fundamental cases of intrinsic periodicity in 3D. For the 2D perception of 3D intrinsic periodicity, the observer's viewpoint can be somewhere in the continuous range between two viewpoint extremes. Ultimately, we distinguish $18$ fundamental cases for the 2D perception of 3D intrinsic periodic motion.

The contributions of our work are the following. (1)~Starting from the first principles of 3D periodicity and its perception in 2D, we derive $18$ fundamentally different cases of repetitive perception. (2)~To estimate repetition in video under realistic circumstances, we compute a diverse flow-based representation over the motion foreground segmentation. Our method uses wavelets to handle non-stationary motion and automatically selects the most discriminative signal based on self-estimated quality assessment. (3)~Extending beyond the video dataset of \cite{levy2015live}, we propose the new QUVA Repetition dataset for repetition estimation, that is more realistic and challenging by lifting the static and stationary assumptions. (4)~We evaluate on the task of repetition counting and show that our method outperforms the deep learning-based state-of-the-art \cite{levy2015live} on the new dataset.


\section{Related Work}
\label{sec:related-work}

Existing approaches for repetition estimation in video typically represent video as one-dimensional signals that preserve the repetitive structure of the motion. Then, frequency information is extracted by Fourier analysis \cite{azy2008segmentation,cutler2000robust,pogalin2008visual,tsai1994cyclic}, peak detection \cite{thangali2005periodic} or singular value decomposition \cite{chetverikov2006motion}. Pogalin \etal \cite{pogalin2008visual} estimate the frequency of motion in video by tracking an object, performing principal component analysis over the tracked regions and employing the Fourier-based periodogram. However, methods relying on Fourier-analysis for periodic motion are unable, nor intended, to handle non-stationary motion as is ubiquitous in the real world.

Briassouli \& Ahuja \cite{briassouli2007extraction} employ time-frequency analysis using the Short Time Fourier Transform for dealing with multiple periodic motions. In \cite{burghouts2006quasi}, the authors propose a spatiotemporal filter bank for estimating repetition in video. Their filters work online and are effective when tuned correctly. However, we question its practical use, as their experiment are limited to stationary motion and the filter bank requires manual tuning. We also use a time-frequency decomposition of signals from video, but concentrate on handling non-stationary repetition. Instead of using the Short-Time Fourier Transform, we adopt the continuous wavelet transform to achieve better resolution \cite{rioul1991wavelets}.

The studies on periodic motion by \cite{davis2000categorical,pogalin2008visual,seitz1997view} have encouraged us to reconsider visual repetition. Pogalin \etal \cite{pogalin2008visual} identify four visually periodic motion types (translation, rotation, deformation and intensity variation) supplemented with three cases of motion continuity (oscillating, constant and intermittent) in the 2D field of view. In this work, we argue that the 3D flow field is the right starting point to derive the foundations of repetition. From the 3D flow field and the differential operators acting on it, we derive three motion types and three motion continuities that organize into a $3\times3$ Cartesian table. Moreover, the projection of 3D periodicity on 2D perception has to consider the viewpoint. What follows are $18$ fundamentally different cases of 2D repetitive perception from 3D periodicity.

Levy \& Wolf \cite{levy2015live} introduce a convolutional neural network for estimating repetition by counting in live video. Their network is trained to predict the motion period on synthetic video sequences in which moving squares exhibit periodic motion of four motion types from \cite{pogalin2008visual}. At test time, the method takes a stack of video frames, computes a region of interest by motion thresholding, and forwards the frame crops through the network to classify the motion period. The system is evaluated on the task of repetition counting and shows near-perfect performance on their \ytsegments{} dataset. The $100$ videos are a good initial set of examples but as the majority of videos have static viewpoint and exhibit stationary periodic repetitions, we propose a new dataset. Our dataset better reflects reality by including more non-static and non-stationary examples. Similar to Levy \& Wolf, we also evaluate repetition estimation by counting.


\begin{figure*}

\centering
    \begin{subfigure}[b]{0.40\textwidth}
        \includegraphics[width=\textwidth,trim={0 7.6cm 17cm 0},clip]{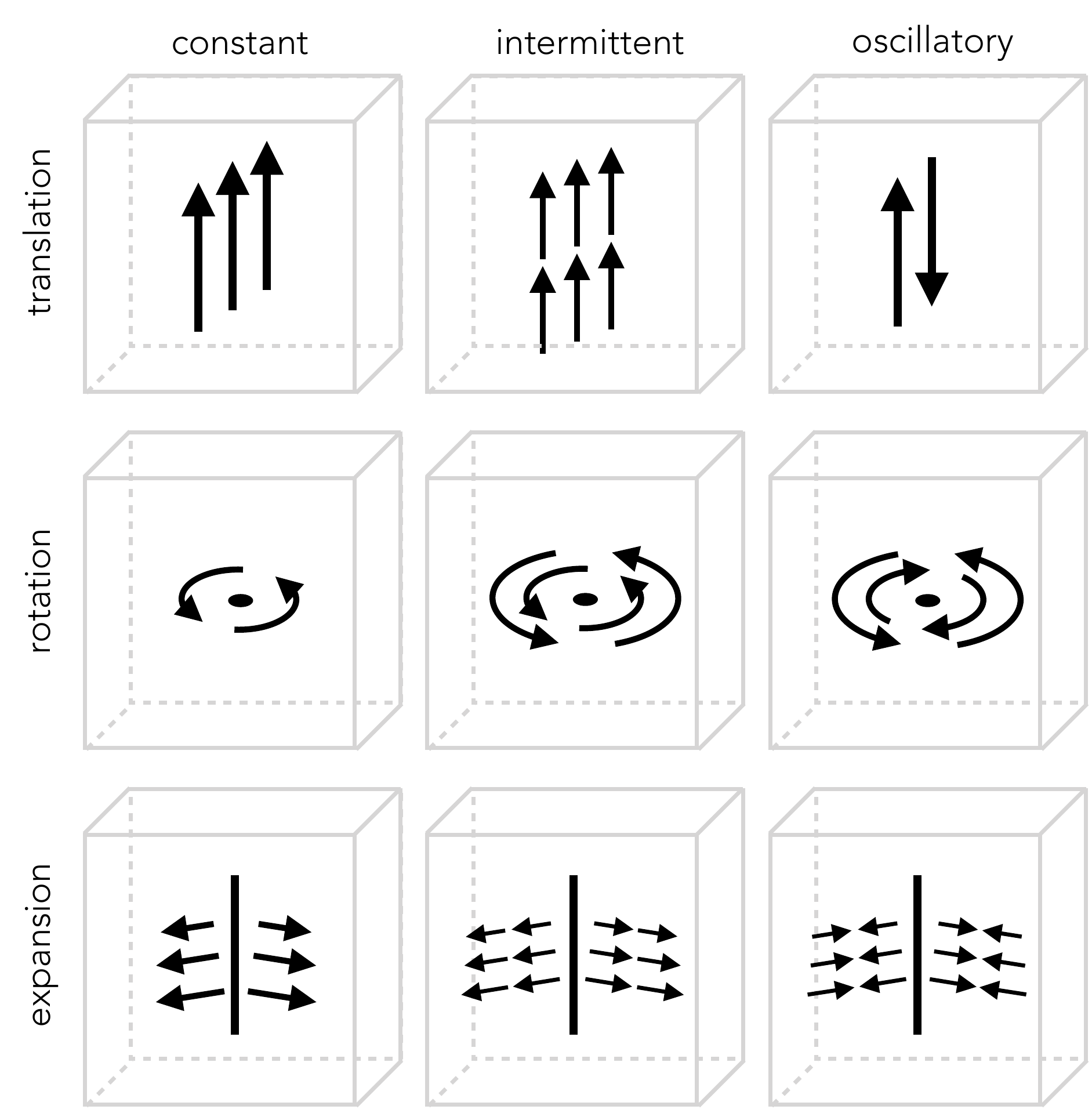}
        \caption{Flow Abstractions in 3D}
        \label{fig:3x3_cubes}
    \end{subfigure}
    \quad
    \begin{subfigure}[b]{0.40\textwidth}
        \includegraphics[width=\textwidth,trim={0 7.6cm 17cm 0},clip]{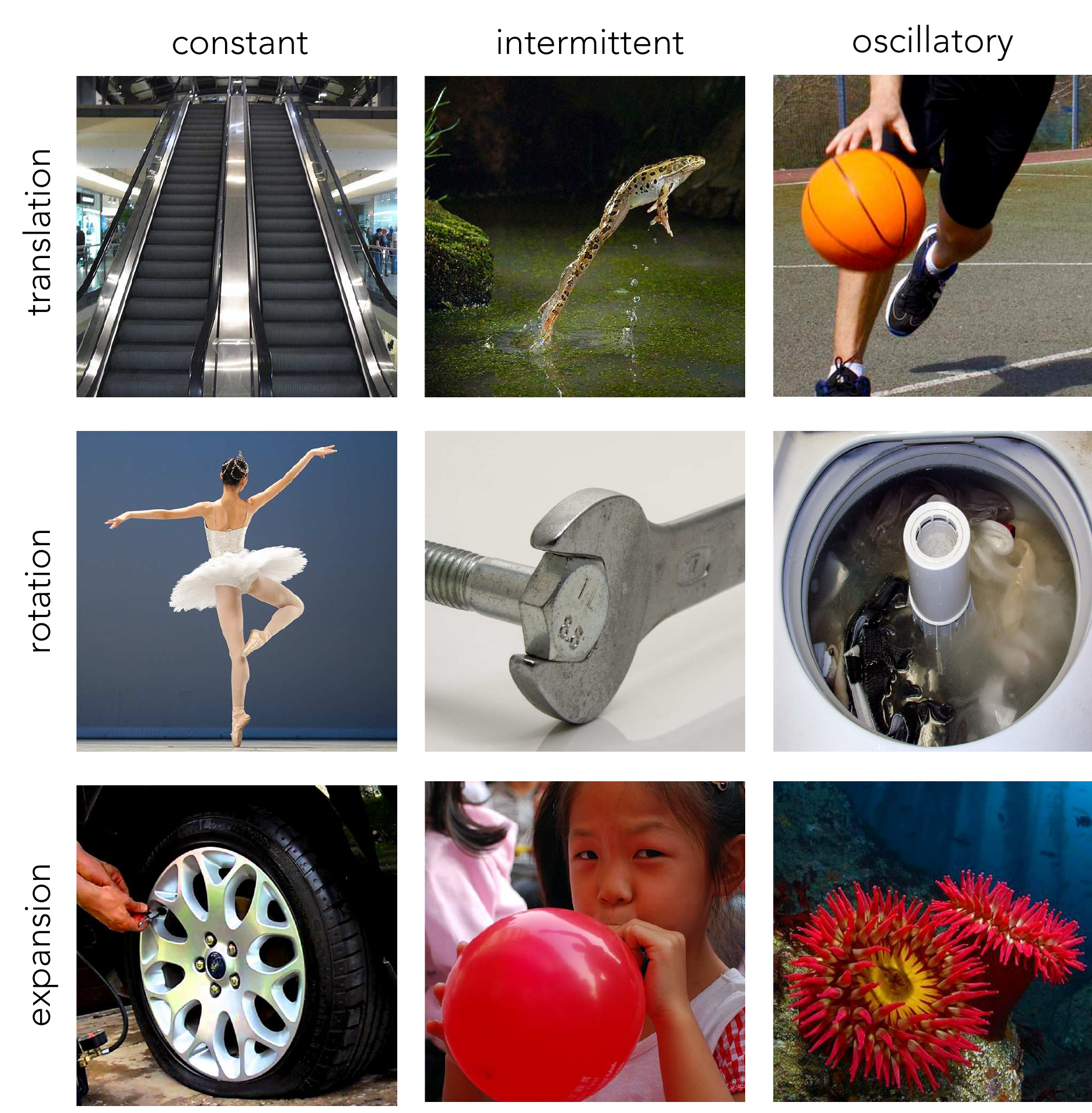}
        \caption{Examples in Real Life}
        \label{fig:3x3_examples}
    \end{subfigure}

    \caption{$3\times 3$ Cartesian table of the \emph{motion type} times the \emph{motion continuity}. Following from the differential operators acting on the flow, these are the basic cases of periodicity in 3D. The examples are: escalator, leaping frog, bouncing ball, pirouette, tightening a bolt, laundry machine, inflating a tire, inflating a balloon and a breathing anemone.} 
    \label{fig:3x3_main_figure}
    \vspace{-5pt}
\end{figure*}

\section{Theory}
\label{sec:recurrence-in-video}

\subsection{3D Intrinsic Periodicity}
\label{subsec:recurrence-in-3d}

In 3D, intrinsic periodicity is defined as the reappearing of the same 3D-flow $\mathcalbf{F}(\*x,t)$ induced by the motion of an object over time. For a moment in time $t$, we denote the flow by $\mathcalbf{F}_t(\*x)$. The 3D-flow field tied to the object is periodic as expressed by $\mathcalbf{F}_t(\*x) = \mathcalbf{F}_{t+T}(\*x + \*S)$, where we exclude for the moment the trivial case that the flow field is constant. The parameter $T$ is the period over time, where $\*S$ is the period, if any, over space.

Let the flow field be given by its directional components: $\mathcalbf{F}_t = (\mathcal{F}_x, \mathcal{F}_y, \mathcal{F}_z)$. From differential geometry, we have the three operators on the flow field:
\begin{align}
  \gradd_t &= \frac{\partial \mathcal{F}_k}{\partial x_j} \widehat{\*e}_j \otimes \widehat{\*e}_k \label{eq:flow-grad} \\
  \divergencee_t &= \frac{\partial \mathcal{F}_x}{\partial x} + \frac{\partial \mathcal{F}_y}{\partial y} + \frac{\partial \mathcal{F}_z}{\partial z} \label{eq:flow-divergence} \\
  \curll_t &=
  \left(\frac{\partial \mathcal{F}_z}{\partial y} - \frac{\partial \mathcal{F}_y}{\partial z}, \frac{\partial \mathcal{F}_x}{\partial z} - \frac{\partial \mathcal{F}_z}{\partial x}, \frac{\partial \mathcal{F}_y}{\partial x} - \frac{\partial \mathcal{F}_x}{\partial y} \right). \label{eq:flow-curl}
\end{align}

Where in Eq.~\eqref{eq:flow-grad} the product $\widehat{\*e}_j \otimes \widehat{\*e}_k$ defines a dyadic tensor, and indices are summed over the $9$ terms by the Einstein convention \cite{spivak1981comprehensive}. The equations define the gradient, divergence and curl of the flow field \cite{schey2005div}. Three basic 3D-motion types emerge depending on the values of divergence and curl as follows:
\vspace{-0.5em}
\begin{align*}
   \text{\emph{translation}:} &\quad \curll_t = \*0, \;\;\; \divergencee_t = 0 \\
   \text{\emph{rotation}:}    &\quad \curll_t \neq \*0, \;\;\; \divergencee_t = 0 \\
   \text{\emph{expansion}:}   &\quad \curll_t = \*0, \;\;\; \divergencee_t \neq 0.
\end{align*}
In practice there may be a mixture types; as we are aiming to handle realistic video, we select the dominant 3D-periodicity in the object's motion whichever is measurable best. In the rare case of counterbalancing expansion and contraction over different axes, it can be that $\divergencee_t = 0$ while being periodic. 

In addition, the motion continuity in 3D can be a source of periodicity. Depending on the type of motion, the motion field needs fulfill one of the following necessary periodic conditions:
\vspace{-1em}
\begin{align*}
   \gradd_t(\*x) &= \gradd_{t+T}(\*x + \*\epsilon) \\
   \curll_t(\*x) &= \curll_{t+T}(\*x + \*\epsilon) \\
   \divergencee_t(\*x) &= \divergencee_{t+T}(\*x + \*\epsilon),
\end{align*}
where $\*\epsilon$ denotes a translation as the object's periodicity may be superposed on translation. For robustness to illumination changes, the measurement of $\gradd_t(\*x)$ is preferred over $\mathcalbf{F}_t$. From these equations three different periodic motion continuities can be distinguished: \emph{constant}, \emph{intermittent} and \emph{oscillating} periodicity. Again, in practice the motion continuity may be a mixture between types.

\subsection{2D Recurrence of 3D Intrinsic Periodicity}
\label{subsec:2d-perception-of-recurrence}

So far we have considered the intrinsic periodicity in 3D. We reserve the term \emph{recurrent} for the 2D observation of the 3D periodicity. Recurrence in the field of view is defined by:
\begin{equation}
  \mathbf{F}_t(\*{\overline{x}}) = \*{F}_{t+T}(\sigma(\*{\overline{x}}+\mathbf{s})),
  \label{eq:flow-2d}
\end{equation}
where $\mathbf{F}_t(\*{\overline{x}})$ is perceived flow in 2D image coordinates $\*{\overline{x}}$, $\mathbf{s}$ is the observed displacement, $T$ is the recurrence and $\sigma$ denotes the observational scale (camera zoom). The underlying principle is that the same period length $T$ will be observed in both 3D and 2D for all cases of intrinsic periodicity. As we perform all measurements within one image, from here on $\*F(\overline{\*{x}})$ implies $\*F_t(\overline{\*{x}})$ where subscript $t$ is omitted for clarity.

In addition, the intrinsic periodicity in 3D does not cover all perceived recurrence in an image sequence. For the trivial cases of constant translation and constant expansion in 3D, perceived recurrence will appear when a repetitive chain of objects (conveyor belt) or a repetitive appearance (checkered balloon) on the object, as given by \autoref{eq:flow-2d}, is aligned with the motion. In such cases, recurrence will also be observed in the field of view. For constant rotation, the restriction is that the appearance cannot be constant over the surface, as no motion, let alone recurrent motion would be observed. In the rotational case, any rotational symmetry in appearance will induce a higher order recurrence as a multiplication of the symmetry and the rotational speed.

For the purpose of recurrence, nine cases organize in a $3\times3$ Cartesian table of basic \emph{motion type} times \emph{motion continuity}, see \autoref{fig:3x3_cubes}. The corresponding examples of these nine cases are given in \autoref{fig:3x3_examples}. This is the list of fundamental cases, where a mixture of types is permitted. In practice, some cases are ubiquitous, while for others it is hard to find examples at all and a mixture of types is rare.

\begin{figure*}[t]
  \centering
  \includegraphics[width=0.98\textwidth]{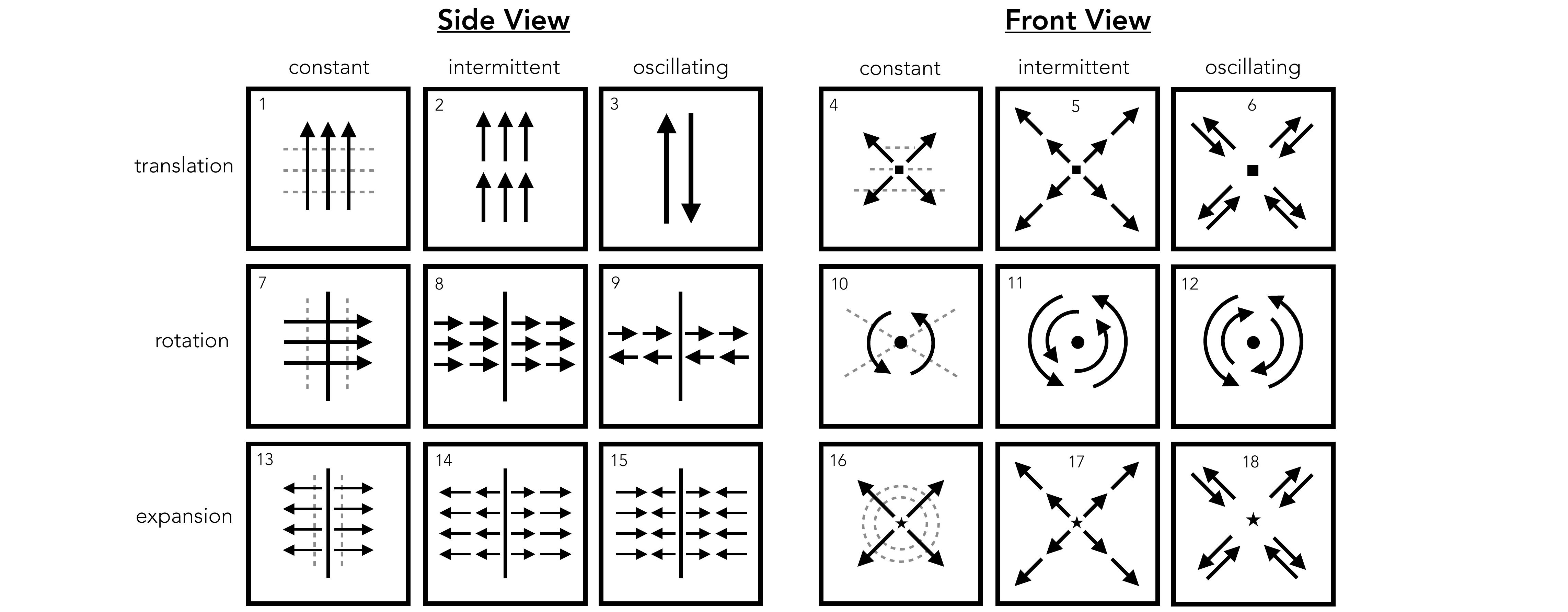}
  \caption{Observed flow: the $18$ fundamental cases for 2D perception of 3D recurrence. The perception follows from the motion pattern ($3\times$), motion continuity ($3\times$) and the viewpoint on the continuous interval between the two extremes: side and front view. $\mathbf{\uparrow}$ denotes flow direction, $\blacksquare$ denotes a vanishing point, $\bullet$ denotes a rotation point, $\bigstar$ denotes expansion point. Dashed grey lines for constant motion indicate the need for texture to perceive recurrence. Pairs $4$-$16$, $5$-$17$ and $6$-$18$ appear similar at first sight but vary in their signal profile. \label{fig:classification-motion-types}}
  \vspace{-10pt}
\end{figure*}

\subsection{The Viewpoint}
\label{subsec:theory-the-viewpoint}

The point of view has a large influence on the perception of the flow field. There are two fundamentally different viewpoints: the \emph{frontal} view and the \emph{side} view:
\begin{align*}
   \text{\emph{frontal view}:} &\quad \text{on the main axis of motion} \\
   \text{\emph{side view}:}    &\quad \text{perpendicular to the main axis of motion}.
\end{align*}

For translation there is one main axis and two perpendicular axes, which are both identical for our purpose. There is no distinction between the two perpendicular views. Similarly, for rotation the two perpendicular cases are also indistinguishable. For expansion there are one, two or three axes of expansion, again leaving us with the frontal case and the perpendicular case as the two fundamental cases. Consequently, for all cases considered, a distinction between frontal view and side view is sufficient. As a result, the perceived recurrence is summarized between the two extreme viewpoints, which results in the Cartesian product of two times nine basic cases as summarized in \autoref{fig:classification-motion-types}. The two views are the end of a continuous range of viewpoints. An actual viewpoint will be somewhere in between the frontal view and the side view, most of the time. This leaves the flow field asymmetrical or skewed, either in gradient, curl or divergence. As long as the signal can be measured this will not affect the recurrent nature of the signal.

\subsection{Non-Static Repetition}
\label{subsec:moving-viewpoint}

So far we have assumed a static camera position. In particular with recurrent motion (1) the camera may move itself because the camera is mounted on the moving object itself, or (2) the camera is following the target of interest, or (3) the camera is in motion independent of the motion of the object. For the first two cases, the camera motion reflects the periodic dynamics of the object's motion. The flow field may be outside the object, but otherwise it displays a complementary pattern in the flow field.

Only the third case demands removal of the camera motion prior to the repetitive motion analysis. In practice, this situation occurs frequently. Therefore, particular attention needs to be paid to camera motion independent of the target’s motion. When due to the camera motion, the viewpoint changes from frontal to side view, the analysis will be inevitably hard. \autoref{fig:classification-motion-types} illustrates the dramatic changes in the flow field when the camera changes from one extreme viewpoint (side) to the other (frontal), or vice versa.

In addition, even when object motion and camera are both static, for none of the intrinsic motion types (translation, rotation, expansion), a point on the object will be at the same position in the camera field all the time. Under the double static condition, a point will just return to the same point on the camera field. As the intermediate points on the object or background have an arbitrary albedo and radiate an arbitrary luminance, no sinusoidal signal will result in general. This is noteworthy as all previous work \cite{cutler2000robust,liu1998finding,pogalin2008visual} implicitly assumed such a signal by considering the Fourier transform or variants.

\begin{figure*}[t]
  \centering
  \includegraphics[width=\textwidth,trim={0 13.6cm 0 0},clip]{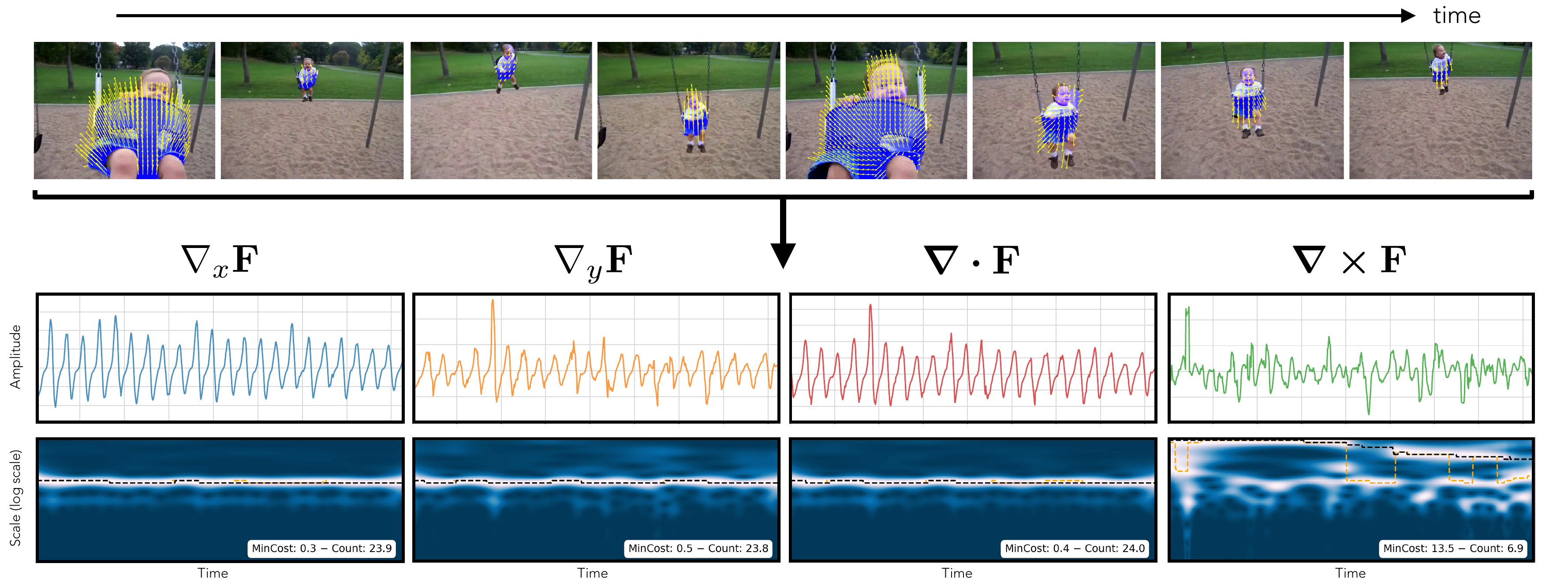}
  \caption{Overview of our method by illustration of an example. First we segment the foreground motion (top row, blue segments) followed by optical flow computation (yellow arrows), then we extract zeroth- and first-order flow signals (4 out of 6 shown) and finally decompose them into a time-frequency spectrum using the continuous wavelet transform (bottom). In the bottom row, the dashed black lines denote the min-cost path whereas the orange lines indicate the maximum power path for counting by integration. Note that for this \emph{oscillatory translation frontal view} case, $\nabla_x F_x$, $\nabla_y F_y$ and $\nablabf \cdot \*F$ give a good signal, as expected, whereas $\nablabf \boldsymbol{\times} \*F$ gives a poor and dispersed signal with heavy cost. \label{fig:system_overview}}
  \vspace{-10pt}
\end{figure*}

\subsection{Non-Stationary Repetition}
\label{subsec:recurrent-signal-characteristics}

A recurrent signal is said to be stationary when the period length is constant over time. In the initial steps of periodicity analysis, it was assumed the periodic signal was near-stationary. In practice, we have observed that stationary repetitive signals are relatively rare. Decay in frequency or accelerating motion are common in realistic video. Therefore, in contrast to \cite{cutler2000robust,pogalin2008visual} we do not assume stationarity, making the method more robust to acceleration. We will employ local wavelets in response to the anticipated signals.


\section{Repetition Estimation}
\label{sec:methods}

Our method for repetition estimation follows a three-stage approach (\autoref{fig:system_overview}). First, we localize the target instance in the scene, then we represent the target by a set of time-varying signals and finally we perform time-frequency decomposition to estimate repetition and select the most discriminative signal.

\vspace{5pt}
\noindent \textbf{Signals from Video}. To deal with camera motion and to handle the wide variety in repetitions, we construct a diverse set of time-varying flow-based signals that we compute over the motion foreground segmentation. Specifically, we measure the average-pooled flow field $\*F = (F_{x}, F_{y})$ and the differentials of the flow. We estimate $\grad$ by measuring $\nabla_x F_x$ and $\nabla_y F_y$. All the differentials of the flow field are computed using Gaussian derivative filters with a large filter size to obtain a global measurement over the foreground segmentation. The final measurement is the average-pooled value over a small radius around the object's center. The differential operators of the flow field comprise four different measurements (as the curl has only one direction perpendicular to the screen), whereas there are two zeroth-order flow signals. In total these amount to six different signals.

For the cases of oscillating and intermittent motion observed from the side, $\grad$ will deliver the strongest repetitive signal. The flow field $\*F$ will convey a stronger repetitive signal for the cases of constant motion appearance. In practice, it may be hard to select the most discriminative signal, to which we return at the end of this section.

\vspace{5pt}
\noindent \textbf{Time-Frequency Decomposition.} Given a discrete signal $h_n$ for timesteps $n = 1, \ldots, N-1$ sampled at equally spaced intervals $\delta t$. Let $\psi_0(\eta)$ be some admissible wavelet function, depending on the non-dimensional time parameter $\eta$. The continuous wavelet transform \cite{grossmann1984decomposition} is defined as the convolution of $h_n$ with a ``daughter'' wavelet generated by scaling and translating the wavelet function $\psi_0(\eta)$:
\begin{equation}
\label{eq:continuous-wavelet-transform}
W_n(s) = \sum_{n'=0}^{N-1} h_{n'} \psi^*\left[\frac{(n'-n)\delta t}{s} \right],
\end{equation}
where the asterisk represents the complex conjugate. By varying time parameter $n$ and the scale parameter $s$, the wavelet transform can generate a time-scale representation describing how the amplitude of the signal changes with time and scale. While formally a time-scale representation, it can also be considered a time-frequency representation since the wavelet scale is directly related to the Fourier frequency \cite{torrence1998practical}. We use the Morlet wavelet, a complex exponential carrier modulated by a Gaussian envelope:
\begin{equation}
\label{eq:morlet-wavelet}
\psi_0(\eta) = \pi^{-1/4} e^{i \omega_0 \eta} e^{\eta^2 / 2}.
\end{equation}
Since the Morlet wavelet is complex, the wavelet transform $W_n(s)$ is also complex. Therefore, it is useful to define the wavelet power spectrum or scalogram as $|W_n(s)|^2$ representing the time-frequency localized energy. The 2D representation can reveal the signal's non-stationary repetitive dynamics. Once the wavelet is chosen, what remains is defining the resolution of the time-frequency spectrum $|W_n(s)|^2$ by specifying scales $s$. In practice, a logarithmic scaling is effective \cite{torrence1998practical}: $s_j = s_02^{j \delta j}$ with $j = 0,1,\ldots,J$. The smallest measurable scale $s_0$ and the number of scales $J$ determine the range of the frequency resolution.

To estimate non-stationary repetitions in a given video, we decompose the six signals into a time-frequency spectrum using the continuous wavelet transform. What follows are six 2D time-frequency representations that enable further analysis of the repetitive contents of the video.

\vspace{5pt}
\noindent \textbf{Counting.} We assume there is only one dominant repetitive motion observable in the wavelet spectrum; this is reasonable as the foreground motion segmentation encourages temporal consistency. Selecting the modulus maximum from the wavelet spectrum $|W_n(s)|^2$ for every timestep $n$ gives a local frequency measurement of approximately $s^{-1}$ for a Morlet wavelet. Our method integrates local frequencies over time to estimate the repetition count: $\widehat{c} = \sum_{n} \delta t / s_{n}$. For a stationary periodic signal the modulus maximum forms a horizontal ridge through time. We emphasize the ability to count non-stationary signals using our approach since the local frequency may change over time. Therefore, our method is able to deal with accelerations or transient phenomena.

\begin{figure}
  \centering
  \scalebox{1.0}[0.9]{
    \includegraphics[width=\columnwidth,trim={0 12cm 7.8cm 0},clip]{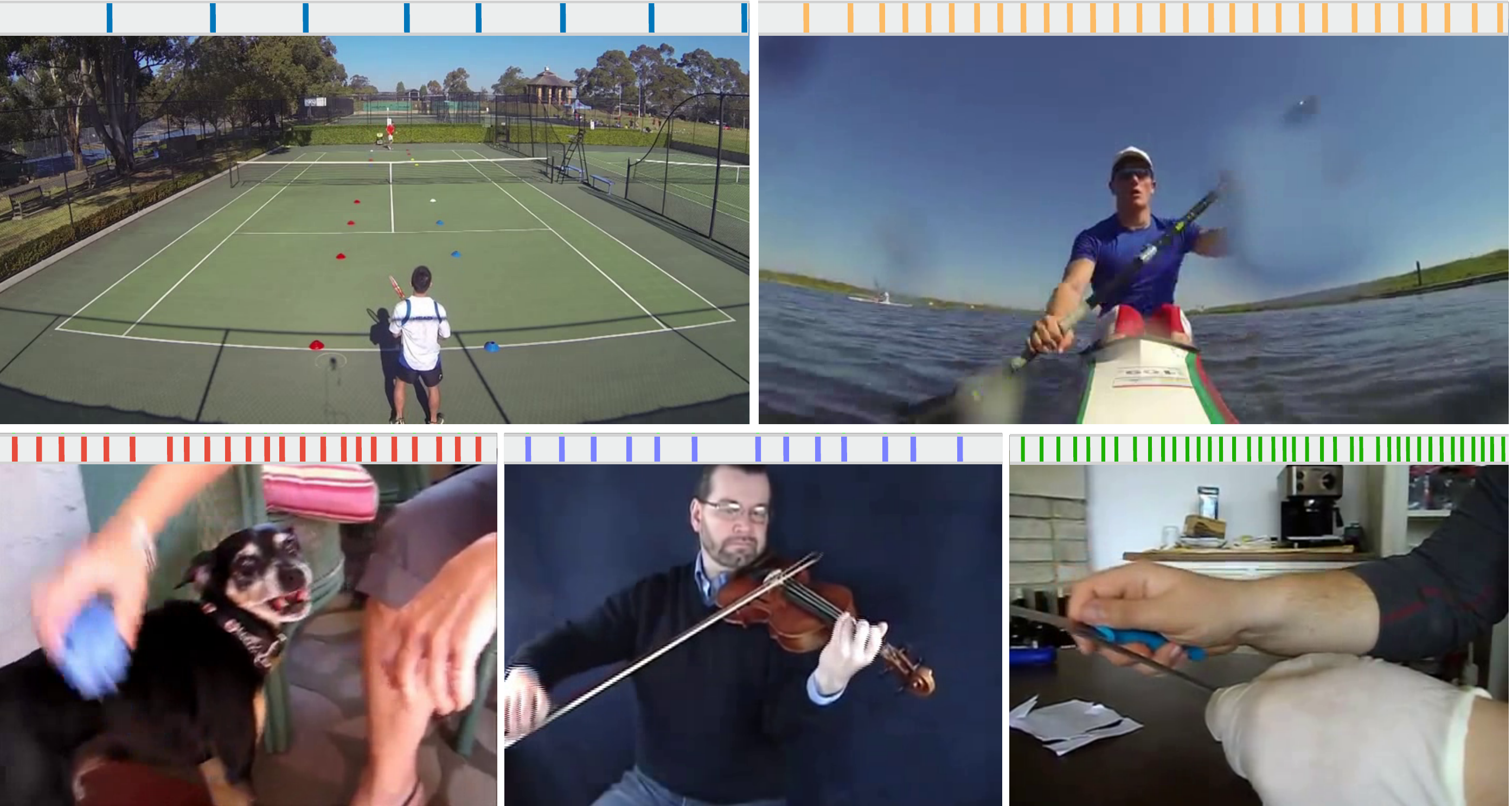}
  }
  \vspace{-15pt}
  \caption{Examples from the \datasetname{} dataset. The timeline with markers illustrate the individual cycle bound annotations, that together determine the final repetition count. Note the diversity in motion appearance and cycle length variability within a video. \label{fig:dataset-examples}}
  \vspace{-12pt}
\end{figure}

\vspace{5pt}
\noindent \textbf{Min-Cost Signal Selection.} The question that remains is selecting the most discriminative signal out of the six. We propose a selection mechanism that prioritizes signals with local regularity in the time-frequency space. Specifically, we adopt a min-cost algorithm for finding the optimal path through the time-frequency space. We turn the wavelet power into a cost surface for optimization by simply inverting it: $1/|W_n(s)|^2$. Traversing over a high-power region translates to low cost. As our goal is to characterize a signal by one cost measure, we run a greedy min-cost pathfinding algorithm to assess the minimum cost required to traverse the spectrum through time. Consequently, the algorithm assigns a lower cost to paths with high local regularity. This is appealing as realistic video signals can be non-stationary but locally smooth. To make a final prediction we select the signal with minimum cost and its corresponding repetition count.


\section{Datasets, Evaluation and Implementation}
\label{sec:dataset-evaluation-implementation}

Motivated by the observation that the \ytsegments{} \cite{levy2015live} dataset for visual repetition estimation is limited in terms of its complexity, we present a new dataset that is more difficult in scene complexity, repetitive appearance and cycle length variation. Our code and data will be made available\footnote{{\footnotesize \url{https://tomrunia.github.io/projects/repetition}}}.

\vspace{5pt}
\noindent \datasetnamebold{} consists of $100$ videos displaying a wide variety of repetitive video dynamics, including swimming, stirring, cutting, combing and music-making. The untrimmed videos are collected from YouTube. We asked two human annotators to label the temporal bounds of each interval containing at least four unambiguous repetitions. We found high inter-agreement between the annotators and keep the $100$ intervals with the highest overlap to increase clarity. Final intervals are obtained by taking the intersection of the two temporal annotations. Next, we ask the annotators to label the repetition count and the temporal bounds of each cycle. \autoref{fig:dataset-examples} shows a few video examples along with their annotation. In \autoref{tab:dataset-statistics} we compare the characteristics of our dataset to the \ytsegments{} \cite{levy2015live}. Our videos have more variability in cycle length, motion appearance, camera motion and background clutter. By increasing difficulty in both scene complexity and temporal dynamics, our dataset represents a more realistic and challenging benchmark for estimating repetition in video.

\begin{table}
\centering
\caption{Dataset statistics of \ytsegments{} \cite{levy2015live} and \datasetname{}. The cycle length variation is the average value of the absolute difference between minimum and maximum cycle length divided by the average cycle length. For this, we annotated all individual cycle bounds in both datasets. The last two rows are also obtained by manual annotation. Note that our dataset is more realistic and challenging in terms of cycle length variability, camera motion and motion complexity. \label{tab:dataset-statistics}}
\vspace{-0.25em}
\resizebox{\columnwidth}{!}{
  \begin{tabular}{lrr}
  \toprule
                     & \ytsegmentsbold{} & \datasetnamebold{} \\
  \midrule
  Number of Videos      & $100$          & $100$             \\
  Duration (s)       & $14.9 \pm 9.8$   & $17.6 \pm 13.3$ \\
  Count Avg.$\pm$ Std.& $10.8 \pm 6.5$   & $12.5 \pm 10.4$   \\
  Count Min/Max      & $4$/$51$         & $4$/$63$               \\
  Cycle Length Variation & $0.22$  & $0.36$            \\
  Camera Motion & $21$ & $53$ \\
  Superposed Translation & $7$ & $27$ \\
  \bottomrule
  \end{tabular}
}
\vspace{-10pt}
\end{table}

\vspace{5pt}
\noindent \textbf{Count Evaluation.}
Given a set of $N$ videos, we evaluate the performance between ground truth count $c_i$ and the count prediction $\widehat{c}_i$ for $i \in \{1,\ldots,N\}$. We report the mean absolute error following prior work \cite{levy2015live}: $\text{MAE} = \frac{1}{N} \sum_{i=1}^N \left|\widehat{c}_i - c_i\right| /c_i$. We also record the off-by-one accuracy (OBOA) or count within-1 accuracy.

\vspace{5pt}
\noindent \textbf{Implementation.} We use the motion segmentation of Papazoglou and Ferrari \cite{papazoglou2013fast}. To account for incorrect segmentation masks we reuse the segmentation of the previous frame if the fraction of foreground pixels is less than $1\%$ of the entire frame. To compute the dense flow field we rely on EpicFlow \cite{revaud2015epicflow}. We compute the divergence and curl by first-order Gaussian derivative filters with a $13\times 13$ filter size. We use a Morlet wavelet with logarithmic scales ($\delta j = 0.125$, $s_0 = 2\delta t$) based on \cite{torrence1998practical} in all experiments. We limit the range of $J$ corresponding to a minimum of four repetitions in the video. Before applying the wavelet transform, we mean filter and linearly detrend the input signals. The mean filter uses a window size of $7$ time steps in all experiments.

\vspace{5pt}
\noindent \textbf{Baselines}. We choose the method of Pogalin \etal \cite{pogalin2008visual} to represent the class of Fourier-based methods for repetition estimation. Our reimplementation uses a more recent object tracker \cite{henriques2012tracker} but is identical otherwise. The tracker is initialized by manually drawing a box on the first frame. Converting the frequency to a count is trivial using the video length and frame rate. Additionally, we compare with the deep-learning method of Levy \& Wolf \cite{levy2015live} using their publicly available code and pretrained model without any modifications.


\begin{figure}
  \centering
  \vspace{-10pt}
  \includegraphics[width=\columnwidth]{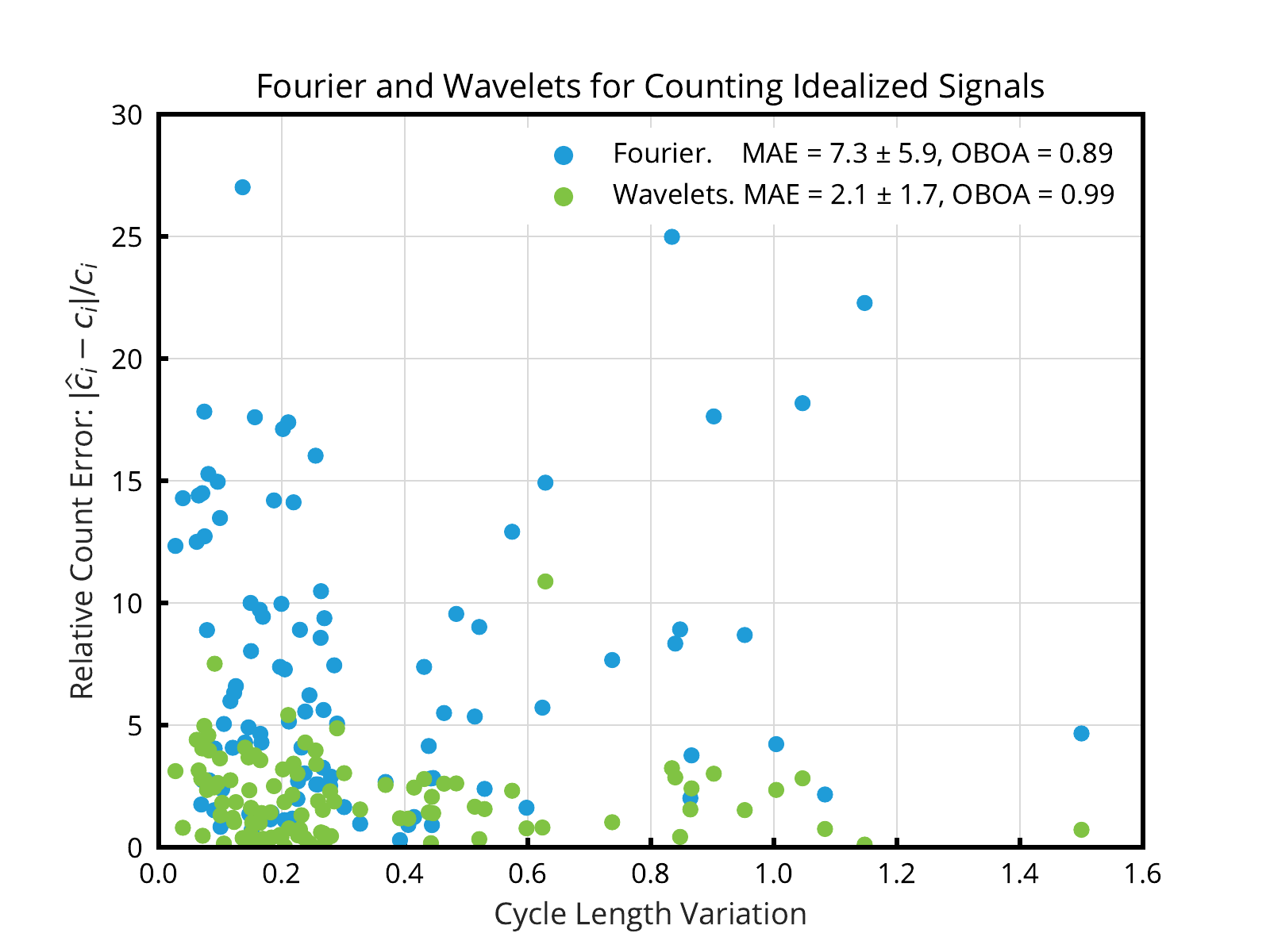}
  \vspace{-15pt}
  \caption{Fourier- versus wavelet-based repetition counting on idealized signals for the videos from the \datasetname{} dataset. Our wavelet-based method outperforms a Fourier-based baseline for $83$ out of $100$ videos. High cycle length variation results in notable error for Fourier measurements, whereas the time-localized wavelets are less sensitive to non-stationary repetition.
  \label{fig:scatter-fourier-wavelets-comparison}}
  \vspace{-10pt}
\end{figure}

\section{Experiments}
\label{sec:experiments}

\subsection{Fourier versus Wavelets}
\label{subsec:experiments-non-statationary}

\noindent \textbf{Setup.} We first compare the Fourier-based periodogram with a wavelet-based time-frequency representation for counting the number of repetitions in each signal. To assess this, we generate idealized signals by plotting sinusoidals through the individual cycle bound annotations for every video in our \datasetname{} dataset. From the periodogram we detect the maximum peak and convert its corresponding frequency to a count using the video's duration.

\vspace{5pt}
\noindent \textbf{Results.} From the results in \autoref{fig:scatter-fourier-wavelets-comparison} it is clear that wavelet-based counting outperforms the periodogram on idealized signals. We also add a significant amount of Gaussian noise ($\sigma = 0.5$) to the signals which has a minor negative effect on both methods (data not shown). We observe that increased cycle length variation negatively affects Fourier-based counting. This is expected as it globally measures frequency and is unable to deal with non-stationarity. As wavelets naturally handle non-stationary repetition they are less sensitive to cycle length variability.

\subsection{Value of Diverse Signals}
\label{subsec:experiments-recurrence-measurements}

\noindent{\textbf{Setup.}} As wavelets prove to be effective for the counting task, we now assess the value of a diverse signal representation. The set of six signals that we verify comprises: $F_x, F_y, \nabla_x F_x, \nabla_y F_y, \divergence, \curl$. These are measured over the foreground segmentation and evaluated for individual performance. Again, we test repetition counting on our \datasetname{} dataset. To obtain a lower-bound on the error, we select the best signal per video in an oracle fashion.

\vspace{8pt}
\noindent{\textbf{Results.}} The results in \autoref{tab:quva-recurrence-individual-signals} reveal that for the wide variability of repetitive appearance there is no one size fits all solution. The individual signals are unable to handle all variety of repetitive appearances by themselves, but their joint diversity results in a good lower-bound. The vertical flow $\*F_y$ is best overall and selected more often than the others by the oracle. We explain this bias towards vertical flow by the observation that our dataset contains many sports videos in which the gravity is often used as opposing force. Repeating this experiment on the \ytsegments{} dataset with oracle signal selection achieves an MAE of $4.2 \pm 5.2$.

\begin{table}
\centering
\caption{Value of diversity in six flow-based signals on videos from our \datasetname{} dataset. The last column denotes how often each signal is selected by the oracle. While the individual signals struggle to obtain good performance by themselves, exploiting their joint diversity is beneficial. \label{tab:quva-recurrence-individual-signals}}
\vspace{-5pt}
\scalebox{0.9}[0.9]{
  \begin{tabular}{lrrcc}
  \toprule
   & MAE & OBOA & \# Selected\\
  \midrule
  $\divergence$     & $44.9 \pm 34.8$ & $0.35$  & $8$ \\
  $\curl$           & $44.9 \pm 34.8$ & $0.42$  & $14$ \\
  $\nabla_x F_x$    & $46.7 \pm 30.8$ & $0.24$  & $12$ \\
  $\nabla_y F_y$    & $42.7 \pm 39.8$ & $0.33$  & $13$ \\
  $F_x$             & $38.3 \pm 31.4$ & $0.40$  & $19$ \\
  $F_y$             & $32.9 \pm 31.4$ & $0.52$  & $34$ \\
  \midrule
  Oracle Best       & $10.5 \pm 15.7$ & $0.81$ & $100$ \\
  \bottomrule
  \end{tabular}
}
\vspace{-6pt}
\end{table}

\subsection{Video Acceleration Sensitivity}
\label{subsec:experiments-halfway-acceleration}

\noindent \textbf{Setup.} In this experiment we examine our method's sensitivity to acceleration by artificially speeding-up videos. Starting from the  \ytsegments{} dataset, we induce significant non-stationarity by artificially accelerating the videos halfway. Specifically, we modify the videos such that after the midpoint frame, the speed is increased by dropping every second frame. What follows are $100$ videos with a $2\times$ acceleration starting halfway. We compare against \cite{levy2015live} which handles non-stationarity by predicting the period of motion in sliding-window fashion over the video. This experiment omits Fourier-based analysis, as by its nature, it will inevitably fail on this task.

\begin{figure}
  \centering
  \vspace{-10pt}
  \includegraphics[width=1.0\columnwidth,trim={0 0 0 0},clip]{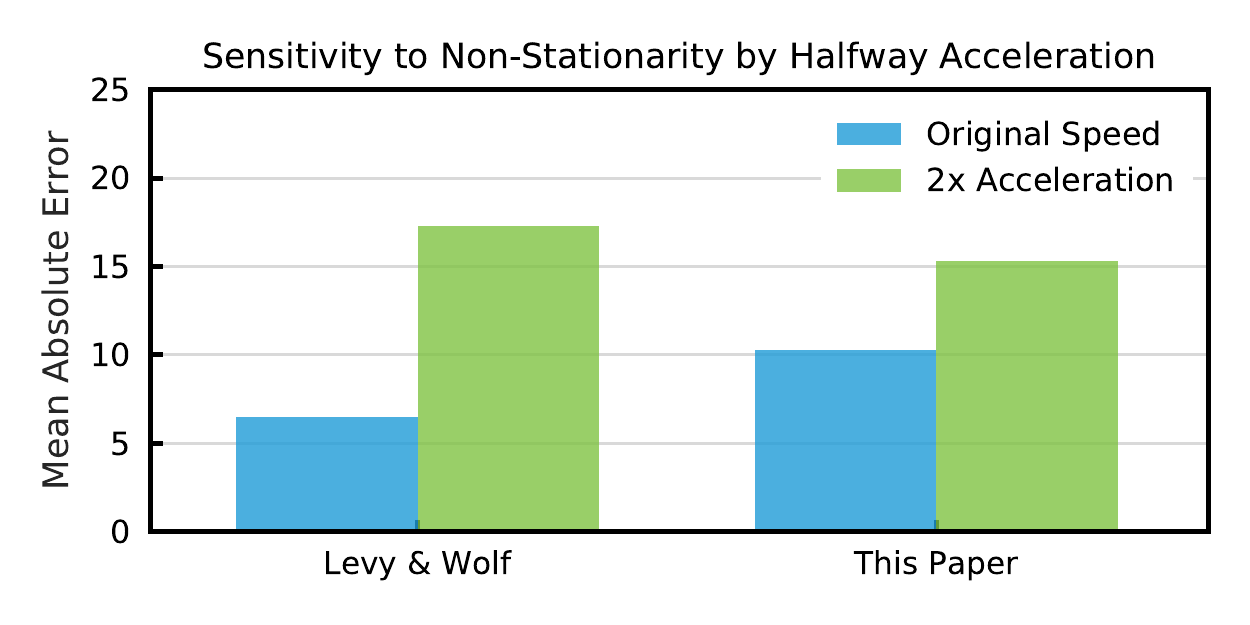}
  \vspace{-0.8cm}
  \caption{The effect of acceleration on the \ytsegments{} dataset. The deep learning method of Levy \& Wolf \cite{levy2015live} has difficulty dealing with non-stationary acceleration, whereas our method suffers less. \label{fig:barchart-acceleration}}
  \vspace{-6pt}
\end{figure}

\vspace{5pt}
\noindent \textbf{Results.} \autoref{fig:barchart-acceleration} presents the MAE in both original and accelerated setting. On their own dataset, the system of Levy \& Wolf \cite{levy2015live} excels. Acceleration changes the results as our method suffers less and obtains a lower MAE on the accelerated videos. This reveals their sensitivity to acceleration, whereas our method deteriorates less.

\subsection{Comparison State-of-the-Art}
\label{subsec:experiments-repetition-counting}

\noindent \textbf{Setup.} We carry out a full count comparison with the methods of Pogalin \etal \cite{pogalin2008visual} and Levy \& Wolf \cite{levy2015live} on both datasets. Our method uses fixed parameters in all cases and utilizes the min-cost signal selection algorithm to pick the most discriminative signal.

\begin{table}[b!]
    \centering
    \vspace{-10pt}
    \caption{Comparison with the state-of-the-art on repetition counting for \ytsegments{} and \datasetname{}. The deep learning-based method of Levy \& Wolf achieves good results on their own dataset of relatively clean videos. On the more realistic and challenging \datasetname{} dataset, our method improves considerably over existing work, be it based on Fourier or deep learning.
    }
    \label{tab:count-evaluation-results}
    \scalebox{0.80}{
    \begin{tabular}{lrrrrrr}
        \toprule
        & \multicolumn{2}{c}{\ytsegmentsbold{} \cite{levy2015live}} & \multicolumn{2}{c}{\datasetnamebold{}} \\
        \cmidrule{2-3} \cmidrule{4-5}
        & MAE $\downarrow$ & OBOA $\uparrow$ & MAE $\downarrow$ & OBOA $\uparrow$ \\
        \midrule
        Pogalin \etal \cite{pogalin2008visual} & $21.9 \pm 30.1$ & $0.68$ & $38.5 \pm 37.6$ & $0.49$ \\
        Levy \& Wolf \cite{levy2015live}       & $\mathbf{6.5 \pm \phantom{0}9.2}$ & $\mathbf{0.90}$ & $48.2 \pm 61.5$ & $0.45$ \\
        This paper & $10.3 \pm 19.8$ & $0.89$ & $\mathbf{23.2 \pm 34.4}$ & $\mathbf{0.62}$ \\
        \bottomrule
    \end{tabular}
}
\vspace{-8pt}
\end{table}

\begin{figure}[b!]
  \vspace{-10pt}
  \centering
  \scalebox{1.0}[0.9]{
    \begin{subfigure}[b]{\columnwidth}
      \includegraphics[width=\textwidth,trim={0 16.75cm 8.4cm 0},clip]{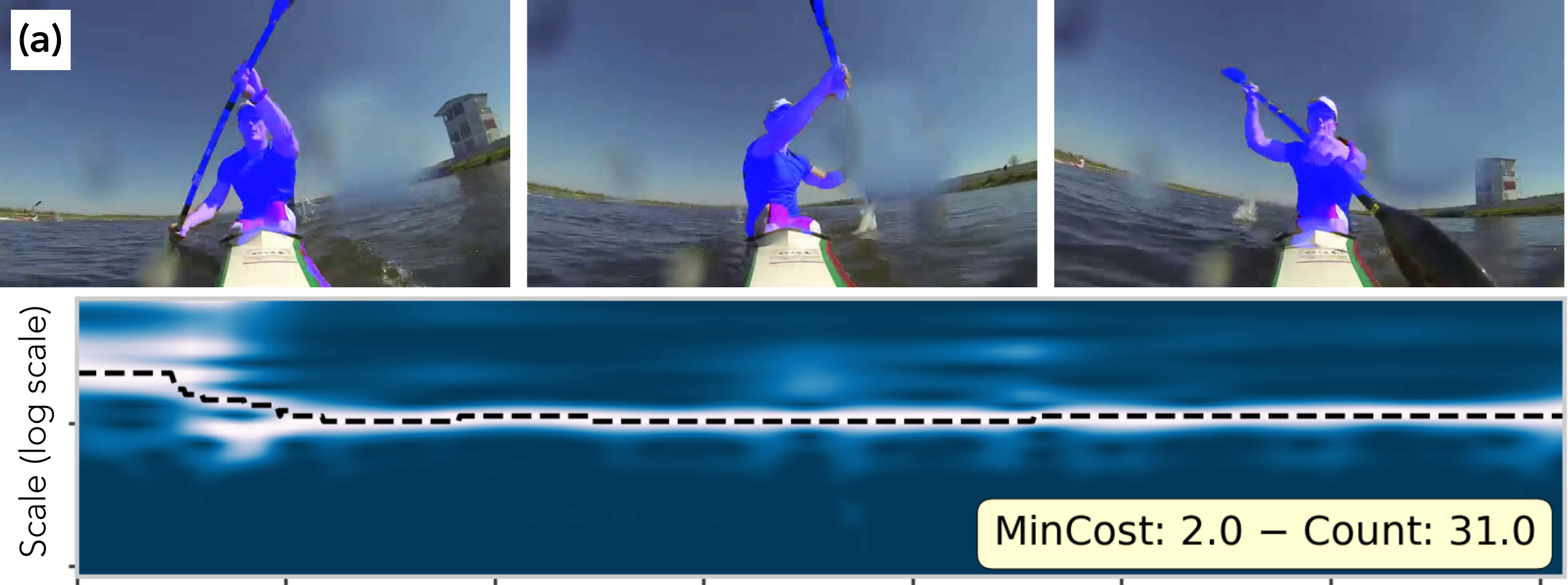}
    \end{subfigure}
  }
  \\[5pt]
  \scalebox{1.0}[0.9]{
    \begin{subfigure}[b]{\columnwidth}
      \includegraphics[width=\textwidth,trim={0 16.75cm 8.4cm 0},clip]{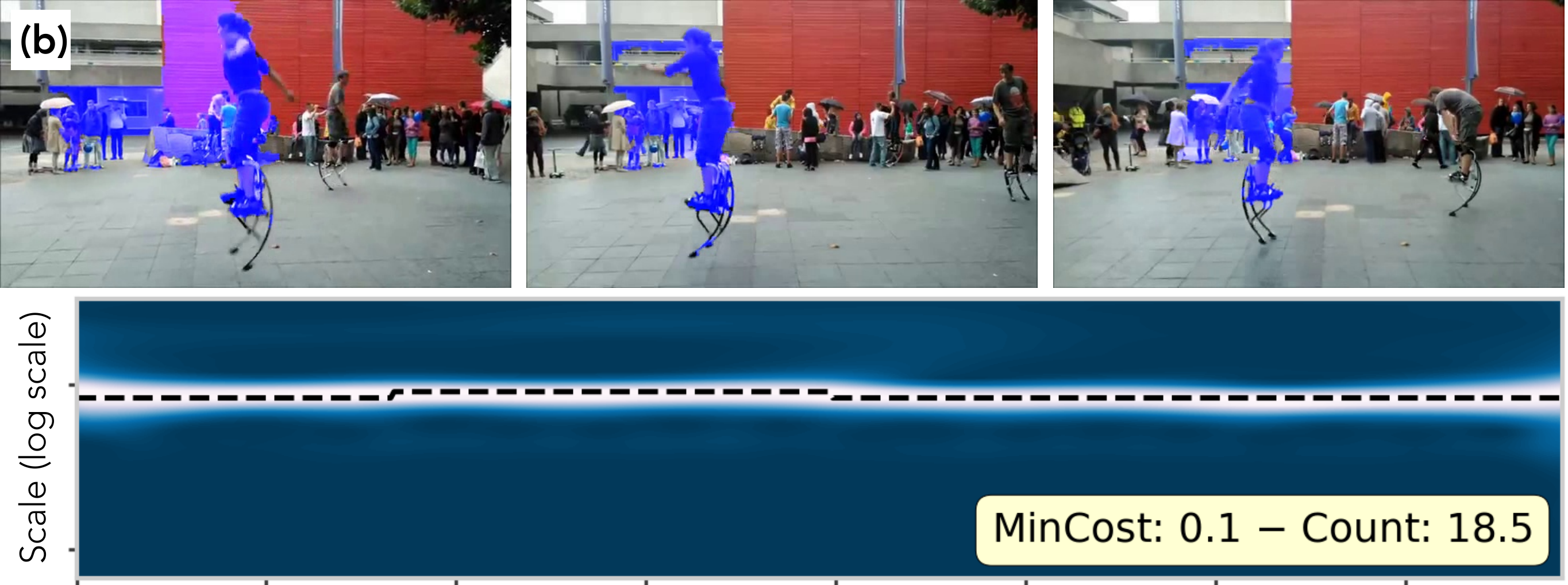}
    \end{subfigure}
  }
  \\[5pt]
  \scalebox{1.0}[0.9]{
    \begin{subfigure}[b]{\columnwidth}
      \includegraphics[width=\textwidth,trim={0 15.9cm 8.4cm 0},clip]{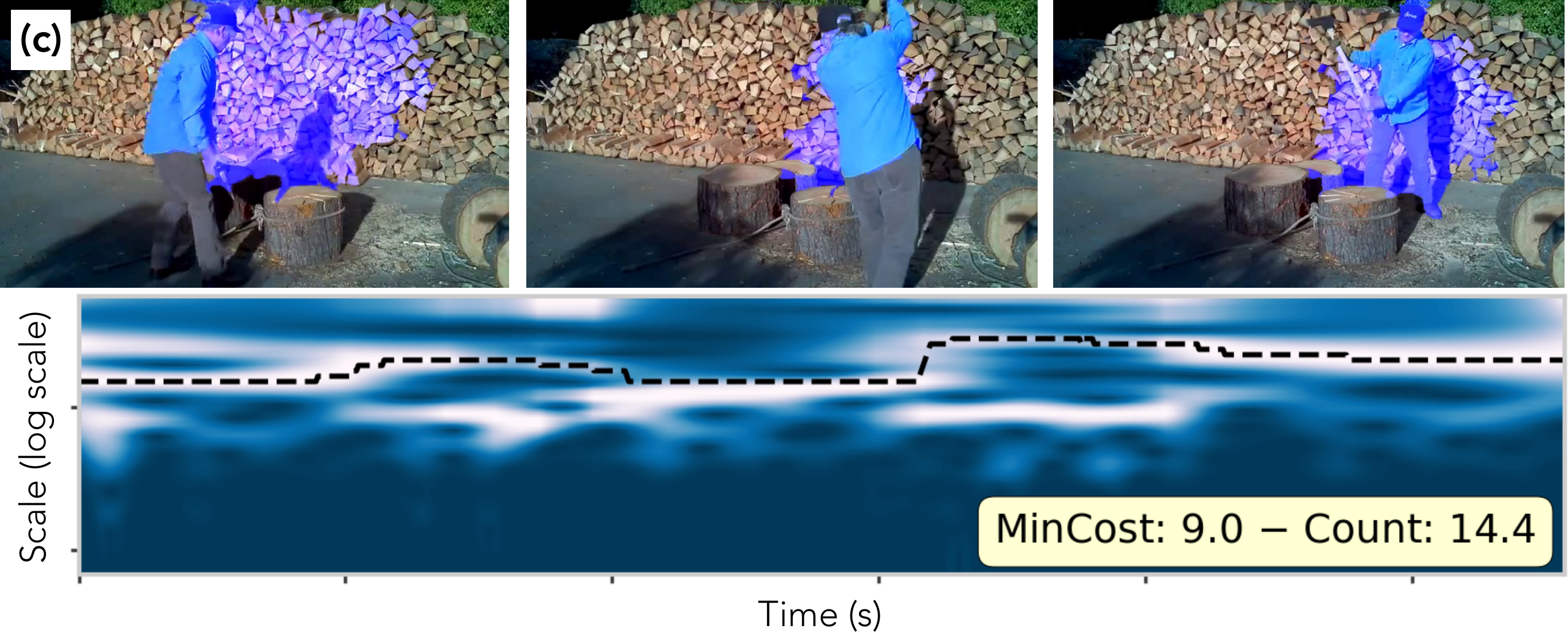}
    \end{subfigure}
  }
  \vspace{-8pt}
  \caption{Results of our method for $3$ video examples. \textbf{(a)} The rower accelerates in the beginning of the video, which appears in the wavelet spectrum of signal $F_x$. Integrating over the max power path results in a repetition count of $31$ whereas the true count is $30$. Our method effectively handles the acceleration. \textbf{(b)} Stationary periodic motion superposed on translation. The video's repetitive nature is evident from the $F_y$ signal. We predict a repetition count of $18.5$ whereas the true count is $18$. \textbf{(c)} Change of viewpoint from \emph{side} to \emph{front} makes this video inevitably hard. Our method is unable to extract a good signal from the video. Note the partial continuity in the spectrum for $\curl$ but distorted by the viewpoint changes. Our method predicts a repetition count of $14.4$ whereas the true count is $16$. \label{fig:final-examples}}
\end{figure}

\vspace{5pt}
\noindent \textbf{Results.} The outcome of the final experiment is presented in \autoref{tab:count-evaluation-results}. For the \ytsegments{} dataset, the method of \cite{levy2015live} performs best with an MAE of $6.5$, where our method scores $10.3$, better than the Fourier-based approach of \cite{pogalin2008visual}. The results change when considering the more realistic and challenging \datasetname{} dataset. The method of \cite{levy2015live} performs the worst, with an MAE of $48.2$, which we attribute to the fact that their network only considers four motion types during training. The Fourier-based method of \cite{pogalin2008visual} scores an MAE of $38.5$, whereas we obtain an error of $23.2$. Overall our method is better able to handle the non-static and non-stationary video characteristics in our \datasetname{} dataset while still performing reasonably well on the videos from \ytsegments{}. We highlight three examples of our method in \autoref{fig:final-examples}.

\section{Conclusion}
\label{sec:conclusion}

We have categorized 3D intrinsic periodic motion as translation, rotation or expansion depending on the divergence and curl of the flow field. Analysis of the time-varying flow gradient distinguishes three motion continuities: constant, intermittent or oscillatory. For the 2D perception of 3D periodicity, two viewpoint extremes are considered. What follows is the categorization of $18$ fundamental cases of recurrent perception derived from the differential operators acting on the flow field. The use of the differentials extends beyond theory, as our experiments demonstrate that measuring flow-based signals over the motion foreground segmentation is effective for recurrence estimation in realistic video. We show that our method improves the state-of-the-art and effectively handles complex appearances, camera motion and non-stationarity on a realistic video dataset.

{\small
  \bibliographystyle{ieee}
  \bibliography{bibliography}
}

\end{document}